\documentclass{bmvc2k}
\usepackage{geometry}
\usepackage{epstopdf}
\usepackage{graphicx}
\usepackage{multirow}
\usepackage{makecell}
\usepackage{mwe}
\usepackage{graphbox}
\usepackage{wrapfig}
\usepackage{color}
\usepackage{amssymb}
\usepackage{array,multirow,graphicx}

\title{DeSTNet: Densely Fused Spatial Transformer Networks\footnote{\textbf{Accepted for publication at the 29th British Machine Vision Conference (BMVC 2018)}}}

\newcommand{\bs}[1]{\mathbf{#1}}

\newcommand{\norm}[1]{\left \Vert#1 \right \Vert}

\addauthor{Roberto Annunziata}{roberto.annunziata@onfido.com}{1}
\addauthor{Christos Sagonas}{christos.sagonas@onfido.com}{1}
\addauthor{Jacques Cal\`i}{jacques.cali@onfido.com}{1}

\addinstitution{
 Onfido Research \\
 3 Finsbury Avenue \\
 London, UK\\
}

\runninghead{Annunziata, Sagonas, Cal\`i}{Densely Fused Spatial Transformer Networks}

\def\etal{\emph{et al}\bmvaOneDot}

\begin{document}

\maketitle

\begin{abstract}
Modern Convolutional Neural  Networks (CNN) are extremely powerful on a range of computer vision tasks.
However, their performance may degrade when the data is characterised by large intra-class variability caused
by spatial transformations. The Spatial Transformer Network (STN) is currently the method of choice for providing CNNs the ability to remove those transformations and improve performance in an end-to-end learning framework. In this paper, we propose \textit{Densely Fused Spatial Transformer Network (DeSTNet)}, which, to our best knowledge, is the first dense fusion pattern for combining multiple STNs. Specifically, we show how changing the connectivity pattern of multiple STNs from \textit{sequential} to \textit{dense} leads to more powerful alignment modules. Extensive experiments on three benchmarks namely, MNIST, GTSRB, and IDocDB show that the proposed technique outperforms related state-of-the-art methods (i.e., STNs and CSTNs) both in terms of accuracy and robustness.
\end{abstract}

\section{Introduction}\label{sec:intro}
Recently, significant progress has been made in several real-world computer vision applications, 
including image classification~\cite{krizhevsky2012imagenet,he2016deep}, face recognition~\cite{schroff2015facenet},
object detection and
semantic segmentation~\cite{girshick2015fast, ren2015faster, he2017mask}. These
breakthroughs are attributed to advances of CNNs
~\cite{huang2017densely, he2016deep, simonyan2014very}, as well as the
availability of huge amounts of data~\cite{krizhevsky2012imagenet,
kemelmacher2016megaface} and computational power. In general, performance is adversely 
affected by intra-class variability caused by spatial transformations, such as affine or 
perspective; therefore, achieving invariance to the aforementioned transformations is highly desirable.
CNNs achieve translation equivariance through the use of convolutional layers. However,
the filter response is not in itself transformation invariant.
To compensate for this max-pooling strategies are often applied
~\cite{boureau2010theoretical, krizhevsky2012imagenet}. Pooling is usually
performed on very small regions (e.g., $2\times2$), giving it an 
effective rate of only a few pixels, increasing as we go deeper.
Another technique used to achieve invariance   
is data augmentation~\cite{krizhevsky2012imagenet}. Specifically, a set of known transformations are applied
to training samples. However, this approach 
has the following disadvantages: \textit{(i)} the set of transformations must be defined \textit{a-priori}; and \textit{(ii)} a large number of samples are required, thus reducing training efficiency.

Arguably, one of best known methods used to efficiently increase invariance to geometric 
transformations in CNNs is the Spatial Transformer Network (STN)~\cite{jaderberg2015spatial}.
STN provides an end-to-end learning mechanism that 
can be seamlessly incorporated into a CNN to explicitly learn how to transform the input data to achieve spatial 
invariance. One might look at an STN as an attention mechanism that
manipulates a feature map in a way that the input is simplified for some process downstream, e.g. image classification.
For example, in~\cite{chen2016supervised} an STN was used in a supervised manner in order 
to improve the performance of a face detector. Similarly, a method based on STN for performing
simultaneously face alignment and recognition was introduced in~\cite{zhong2017toward}.
Although the incorporation of the STN within CNNs led to state-of-the-art performance,
its effectiveness could reduce drastically in cases where the
face is heavily deformed (e.g. due to facial expressions). To overcome this issue, Wu~\etal~\cite{wu2017recursive}
proposed multiple STNs linked in a recurrent manner.
One of the main drawbacks when combining multiple STNs can be seen in the boundary pixels. Each STN samples the output image produced
by the previous, thus as the image passes through multiple transforms the quality
of the transformed image deteriorates. In cases where initial bounding boxes are not
of sufficient accuracy, transformed images are heavily affected by the 
boundary effect, shown in~\cite{lin2017inverse}. To overcome this and inspired by the Lucas-Kanade 
algorithm~\cite{lucas1981iterative}, Lin and Lucey~\cite{lin2017inverse}
proposed Compositional STNs (CSTNs) and their recurrent version ICSTNs. CSTNs are made up of an STN variant (henceforth, \textbf{p}-STN), which propagates transformation parameters instead of the transformed images. 
 
\begin{figure}[t]
\label{fig:main_figure}
\centering
\includegraphics[width=0.99\textwidth]{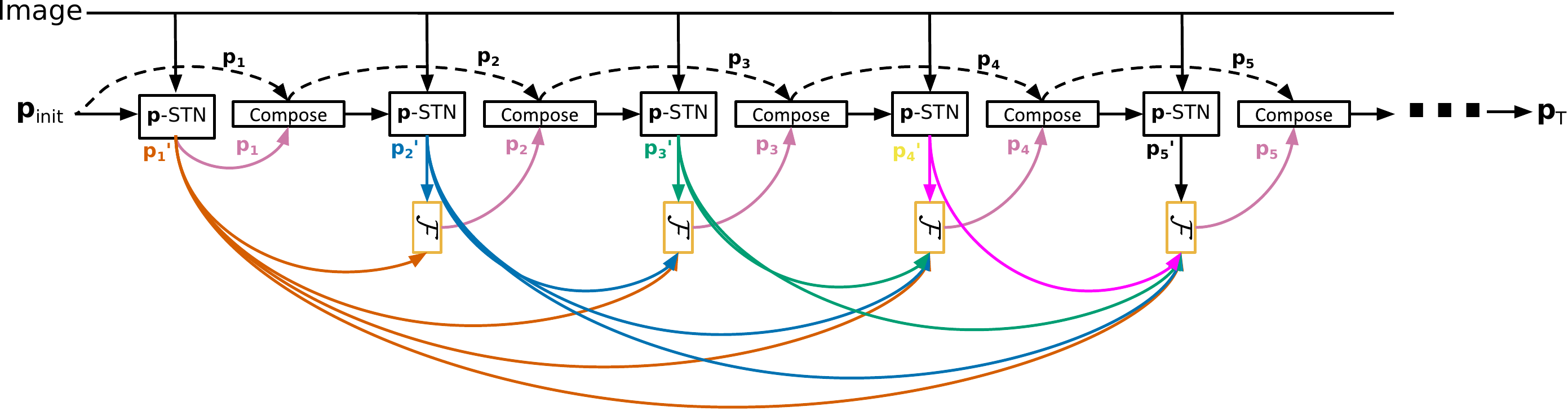} 
\caption{DeSTNet - A stack of Densely fused Spatial Transformer Networks.}
\end{figure}

In this work, building on the success of \textbf{p}-STNs, we present DeSTNet (Fig.~\ref{fig:main_figure}), an 
end-to-end framework designed to increase spatial invariance in CNNs. Firstly, motivated by information theory
principles, we propose a dense fusion connectivity pattern for \textbf{p}-STNs. Secondly, we introduce a novel \textit{expansion-contraction} fusion block for combining the predictions of multiple \textbf{p}-STNs in a \textit{dense} manner. Finally, extensive experimental results on two public benchmarks and a non-public real-world dataset suggest that the proposed DeSTNet outperforms the state-of-the-art
CSTN\cite{lin2017inverse} and the original STN~\cite{jaderberg2015spatial}.
\section{Related Work}\label{sec:related}

Geometric transformations can be mitigated through the use of either \textit{(i)} invariant or equivariant features;
\textit{(ii)} encoding some form of attention mechanism. More traditional
computer vision systems achieved this through the use of hand-crafted
features such as HOG~\cite{dalal2005histograms}, SIFT~\cite{lowe2004distinctive} and SCIRD~\cite{annunziata2015scale,annunziata2016accelerating} that
were designed to be invariant to various transformations. In CNNs
translation equivariance is achieved through convolutions and limited spatial invariance from pooling.

In ~\cite{kanazawa2014locally}, a method for creating scale-invariant CNNs was proposed.
Locally scale-invariant representations are obtained by applying filters at 
multiple scales and locations followed by max-pooling. Rotational
invariance can be achieved by discretely rotating the
filters~\cite{cohen2016group, cohen2016steerable, marcos2016learning} or input images
and feature maps~\cite{oyallon2015deep, laptev2016ti, dieleman2016exploiting}. Recently,
a method for providing continuous rotation robustness was proposed
in~\cite{worrall2017harmonic}. To facilitate the translation invariance property
of CNNs, Henriques and Vedaldi~\cite{henriques2016warped} proposed to transform the
image via a constant warp and then employ a simple convolution. Although, the aforementioned
is very simple and powerful, it requires prior knowledge of the type of transformation as well 
as the location inside the image where it is applied. 

More related to our work are methods that encode an attention or detection
mechanism. Szegedy~\etal ~\cite{szegedy2013deep} introduced a detection system 
as a form of regression within the network to predict object bounding
boxes and classification results simultaneously. Erhan~\etal~\cite{erhan2014scalable}
proposed a saliency-inspired neural network that predicts a set of class-agnostic bounding boxes along
with a likelihood of each box containing the object of interest. A few years later,
He~\etal~\cite{he2017mask} designed a network that performs a number of complementary
tasks: classification, bounding box prediction and object segmentation. The region proposal
network within their model provided a form of learnt attention mechanism.
For a more thorough review of object detection systems we point the reader to
Huang~\etal ~\cite{huang2017speed} who look at speed/accuracy
trade-offs for modern detection systems.

\section{Methodology}\label{sec:method}
Let $\mathcal{D}=\{\bs{I}_1, \bs{I}_2, \ldots, \bs{I}_M\}$ be a set of \textit{M} images and $\{\bs{p}_i\}_{i=1}^M \in \mathbb{R}^n$ ($n=8$ for perspective)\footnote{This initial estimation may simply be an identity.} the initial estimation of the distortion parameters for each image. Our goal is to reduce the intra-class variability due to the perspective transformations
inherently applied to the images during capture. Achieving this goal has the potential to
significantly simplify subsequent tasks, such as classification. To this end, we need to find
the optimal parameters $\{\bs{p}_i^*\}_{i=1}^M$ that warp all the images into a transformation-free space.

\begin{figure}[t]
\begin{tabular}{cc}
\includegraphics[width=0.35\textwidth]{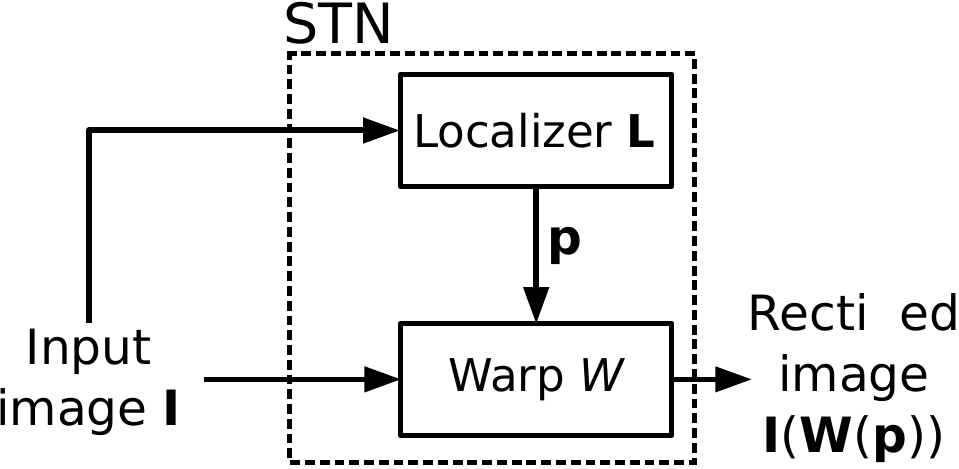} 
& \includegraphics[width=0.6\textwidth]{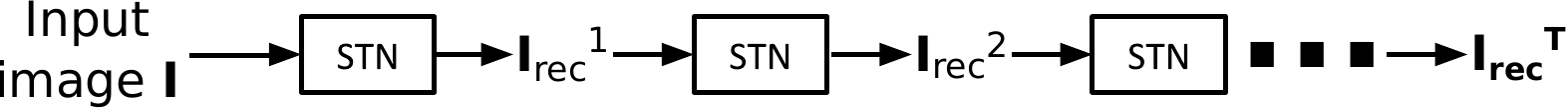} 
\\ (a) & (b)
\end{tabular}
\caption{(a) Spatial Transformer Network (STN)~\cite{jaderberg2015spatial} and (b) stack of STNs.}
\label{fig:stn_figs}
\end{figure}

Arguably, the most notable method for finding the optimal parameters is the STN~\cite{jaderberg2015spatial}. An STN is made up of three components, namely the
\textit{localization network}, the \textit{grid generator} and the \textit{sampler}. The
\textit{localization network} $\bs{L}$ is used to predict transformation parameters
for a given input image $\bs{I}$ and initial parameters $\bs{p}_\text{init}$, i.e. $\bs{p} = \bs{L}(\bs{I}, \bs{p}_{\text{init}})$,
the \textit{grid generator} and \textit{sampler} are  used for warping the image based on the computed
parameters, i.e. $\bs{I}(\mathcal{W}(\bs{p}))$ (Fig.~\ref{fig:stn_figs}(a)). By allowing the network to learn how to warp the input, it is able to gain geometric invariance, thus boosting task performance.
When recovering larger transformations a number of STNs can be stacked or used in combination with a 
recurrent framework (Fig.~\ref{fig:stn_figs}(b)). However, this tends to introduce boundary artifacts
and image quality degradation in the final transformed image, as each STN re-samples from an image that
is the result of multiple warpings. 

To address the aforementioned and inspired by the success of the LK algorithm for image
alignment, Lin and Lucey~\cite{lin2017inverse} proposed compositional STNs (CSTNs). The LK algorithm is
commonly used for alignment problems~\cite{baker2004lucas,matthews2004active} as it approximates the linear relationship between appearance and geometric displacement. Specifically, given two images $\bs{I}_1$, $\bs{I}_2$ that are related by a parametric transformation $\mathcal{W}$, the goal of LK is to find the optimal parameters that minimize the $\ell_2$ norm of the error between the deformed version of $\bs{I}_1$, and $\bs{I}_2$: $\underset{\bs{p}}{min} \norm{\bs{I}_1(\mathcal{W}(\bs{p})) - \bs{I}_2}_{2}^{2}$.
%
Applying first-order Taylor expansion to $\bs{I}_1$, it has been shown that the previous problem can be optimised by an iterative algorithm with the following additive-based update rule: 
\begin{equation}
\label{eq:LK_updates}
\bs{p}_{t+1}  = \bs{p}_{t} + \Delta \bs{p}_t,
\end{equation}
at each iteration $t$. In~\cite{lin2017inverse}, Lin and Lucey introduced the CSTN that predicts the parameters' updates by employing a modified STN, which we refer to as $\bs{p}$-STN, and then compose them as in Eq.~(\ref{eq:LK_updates}).
\begin{figure}[t]
\centering
\begin{tabular}{cc}
\includegraphics[width=0.35\textwidth]{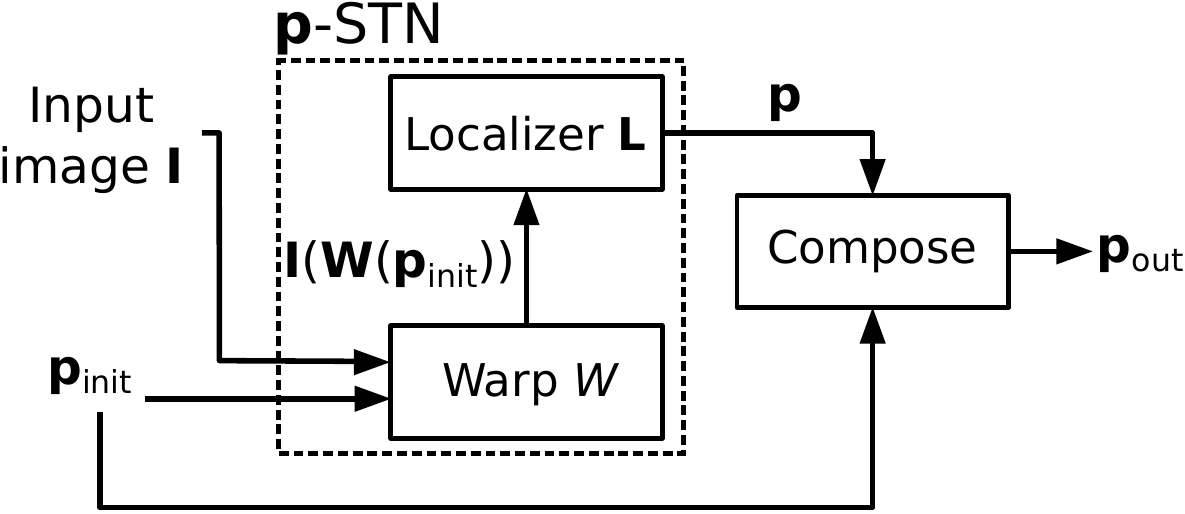} 
& \includegraphics[width=0.60\textwidth]{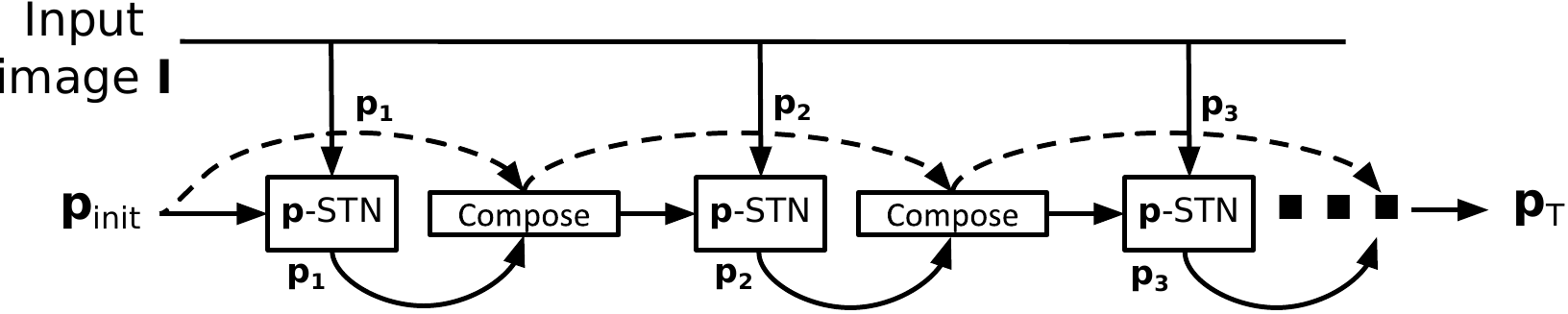} 
\\ (a) & (b)
\end{tabular}
\caption{(a) Compositional STN (CSTN)~\cite{lin2017inverse} and (b) stack of CSTNs.}
\label{fig:lk_stn}
\end{figure}
By incorporating the LK formulation, the resulting CSTN is able to inherit the geometry preserving property of LK. Unlike a stack of STNs that propagates $\textit{warped images}$ to recover large displacements~(Fig.~\ref{fig:stn_figs}(b)), a stack of CSTNs (Fig.~\ref{fig:lk_stn}(b)) propagate the \textit{warp parameters} in a similar fashion to the iterative process used in the LK algorithm.

Here, we extend the CSTN framework to improve the information flow in terms of parameters' updates. In particular, we modify Eq.~(\ref{eq:LK_updates}) and propose the additive-based \textit{dense fusion} update rule:
\begin{equation}
\label{eq:Dense_LK_updates_function}
\bs{p}_{t+1}  = \bs{p}_{t} + f(\Delta \bs{p'}_t, \Delta \bs{p'}_{t-1}, \ldots, \Delta \bs{p'}_1),
\end{equation}
where the parameters' update at iteration $t$, $\Delta \bs{p}_t$, is now a function $f:
\mathbb{R}^{n\times t} \rightarrow \mathbb{R}^n$ of the updates predicted by the $\bs{p}$-STN
at iteration $t$, $\Delta \bs{p'}_t$, and $\it{all}$ the previous ones, $\{\Delta \bs{p'}_i\}_{i=1}^{t-1}$
(Fig.~\ref{fig:main_figure}). Learning the \textit{fusion} function $f(\cdot)$ at each iteration $t$ means learning the posterior distribution $p(\Delta \bs{p}_t|\Delta \bs{p'}_t, \Delta \bs{p'}_{t-1}, \ldots, \Delta \bs{p'}_1)$ for the parameters' update $\Delta \bs{p}_t$. From an information theory perspective, this amounts to predicting $\Delta \bs{p}_t$ with an \textit{uncertainty} measured by the conditional entropy, $\mathcal{H}(\Delta \bs{p}_t|\Delta \bs{p'}_t, \Delta \bs{p'}_{t-1}, \ldots, \Delta \bs{p'}_1)$. We notice that the CSTN update in Eq.~(\ref{eq:LK_updates}) is a special case of Eq.~(\ref{eq:Dense_LK_updates_function}):
\begin{equation}
\label{eq:LK_updates_function}
\bs{p}_{t+1}  = \bs{p}_{t} + f(\Delta \bs{p'}_t),
\end{equation}
where the parameters' update at iteration $t$, $\Delta \bs{p}_t$, is \textit{only} a function 
of the update predicted by the $t^{th}$ regressor ($\Delta \bs{p'}_t$). In fact, no fusion has to be applied (i.e., $f(\cdot)$ is an identity mapping) and $\Delta \bs{p}_t = \Delta \bs{p'}_t$. In other words, the CSTN
learns the 
distribution $p(\Delta \bs{p}_t)$ for the parameters' update $\Delta \bs{p}_t$ at each 
iteration $t$. This amounts to predicting $\Delta \bs{p}_t$ with an 
\textit{uncertainty} measured by the related entropy, $\mathcal{H}(\Delta \bs{p}_t)$. Invoking the well-known \textit{`conditioning reduces entropy'} principle from information theory~\cite{cover2012elements}, it can be shown that $\mathcal{H}(\Delta \bs{p}_t|\Delta \bs{p'}_t, \Delta \bs{p'}_{t-1}, \ldots, \Delta \bs{p'}_1) \leq \mathcal{H}(\Delta \bs{p}_t)$. In other words, the update predictions in the proposed formulation are \textit{upper}-bounded by those made with CSTN in terms of uncertainty. We advocate that this theoretical advantage can translate into better performance.
\begin{figure}[!t]
\centering
\begin{tabular}{cc}
\includegraphics[height=0.19\textheight]{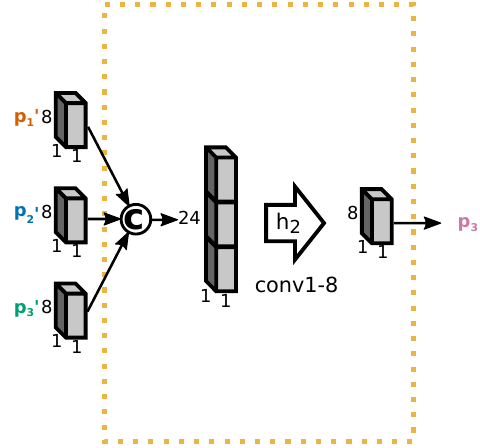} & \includegraphics[height=0.19\textheight]{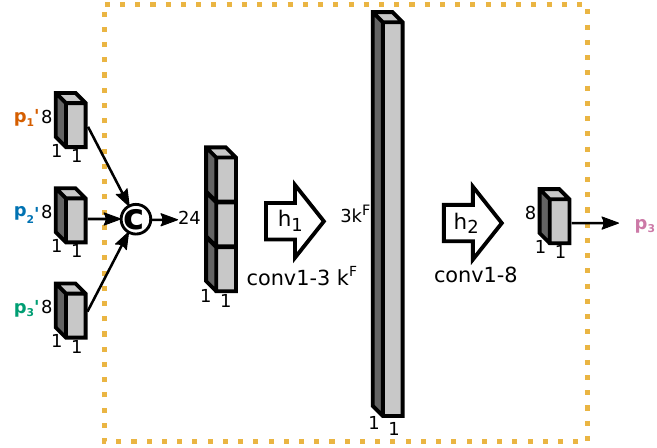} \\
(a) & (b)
\end{tabular}
\caption{Fusion blocks. (a) The bottleneck-based fusion block proposed in~\cite{huang2017densely}. (b) The proposed \textit{expansion-contraction} fusion block used in Figure~\ref{fig:main_figure}.}
\label{fig:fusion}
\end{figure}

Inspired by the recent success of densely connected CNNs~\cite{huang2017densely} and justified by the extension outlined above, we propose an alignment module which we call \textit{DeSTNet} (Densely fused Spatial Transformer Network). DeSTNet consists of a cascade of $\bs{p}$-STNs with a dense fusion connectivity pattern, as shown in Fig.~\ref{fig:main_figure}. The fusion function, implemented by the fusion block $\mathcal{F}$ in Fig.~\ref{fig:main_figure}, is adopted to combine the update predictions of $\textit{all}$ the previous $\bs{p}$-STNs and estimate the best parameters' update at each level $t$. Unlike the fusion blocks adopted in~\cite{huang2017densely} consisting of a single bottleneck layer (Fig.~\ref{fig:fusion}(a)), we advocate the use of an \textit{expansion-contraction} fusion block (Fig.~\ref{fig:fusion}(b)). This solves the fusion task in a high-dimensional space and then maps the result back to the original. Specifically, we concatenate all the previous parameters' updates and project them using a $1 \times 1$ convolution layer with depth $n\times t \times k^F$ (\textit{expansion}), where $n$ is the dimension of the warp parameters $\bs{p}$, $t=1,\ldots, T$ is the level within DeSTNet, and $k^F$ is the \textit{expansion rate}. This is then followed by a $1 \times 1 \times n$ convolution layer (\textit{contraction}), as shown in Fig.~\ref{fig:fusion}(b). We adopt \text{tanh} activations (non-linearities) after each convolutional layer of the fusion block to be able to predict both positive and negative parameter values. It is worth noting that the use of expansion layers is made possible by the relatively low dimension of each individual prediction (i.e., $n = 8$ for perspective warps).

%
%
\section{Experiments}\label{sec:experiments}
In this section, we assess the effectiveness of the proposed DeSTNet in \textit{(i)} adding spatial transformation invariance (up to perspective warps) to CNN-based classification models and ~\textit{(ii)}~planar image alignment. To this end, artificially distorted versions of two widely used datasets, namely the German Traffic Sign Recognition Benchmark (GTSRB)~\cite{stallkamp2011german} and MNIST~\cite{lecun1998mnist} are utilised. Furthermore, we evaluate the performance of DeSTNet on a non-public dataset of official identity documents (IDocDB), which includes substantially larger images (e.g. up to $6,016\times3,910$ pixels) and, more importantly, real perspective transformations. Additional results can be found in supplementary material. 
\subsection{Image Classification}
\label{sec:exp_traffic}
\textbf{Traffic Signs:} We report experimental results on the GTSRB dataset~\cite{stallkamp2011german},
consisting of $39,209$ training and $12,630$ test colour images from $43$ traffic signs taken
under various real-world conditions including motion blur, illumination changes and extremely
low resolution. We adopt the image classification error as a \textit{proxy} measure for
alignment quality. Specifically, we build classification pipelines made up of two components:
an alignment network followed by a classification one (detailed architectures reported in Table~\ref{tbl:ts_mnist_pertrubed}). Both networks are jointly trained with the classification-based
loss using standard back-propagation. At parity of a classification network, a lower
classification error suggests better alignment (i.e., spatial transformation invariance). Following the experimental protocol
in~\cite{lin2017inverse}, we resize images to $s\times s$, $s=36$ pixels and artificially
distort them using a perspective warp. Specifically, the four corners of each image
are independently and randomly scaled with Gaussian noise $\mathcal{N}(0,(\sigma s)^2)$,
then randomly translated with the same noise model. 

In the first experiment, we follow the same setting adopted in~\cite{lin2017inverse} and train all the networks for 
200,000 iterations with a batch of 100 perturbed samples generated on the fly. For DeSTNet, we use 
$\alpha_{\text{clf}}=10^{-2}$ as the learning rate for the classification network and $\alpha_{\text{aln}}=10^{-4}$ for 
the alignment network which is reduced by $10$ after $100,000$ iterations. For the proposed \textit{expansion-contraction }
fusion block we set the expansion rate $k^{F}=256$, as a good trade-off between speed and performance, 
we use dropout with keep probability equal to $S=0.9$. Finally, images of both train and test sets are 
perturbed using $\sigma = 10\%$, corresponding to a maximum perturbation of $3.6$ pixels. 
\begin{table}[t]
\centering
\begin{scriptsize}
\begin{tabular}{c|c|c|c|c}
                         & \multirow{2}{*}{\textbf{Model}} & \multirow{2}{*}{\textbf{Test Error}} & \multicolumn{2}{c}{\textbf{Architecture}} \\ \cline{4-5} 
                         &                        &                             & \textbf{Alignment}       & \textbf{Classifier}      \\ \hline
\multirow{6}{*}{\rotatebox[origin=c]{90}{{\textbf{GTSRB}}}} & CNN                    & $8.29\%$ & \multicolumn{2}{c}{conv$7$-$6$ | conv$7$-$12$ | P | conv$7$-$24$ | FC($200$) | FC($43$)}             \\ \cline{2-5} 
                         & STN & $6.49\%$ & conv$7$-$6$ | conv$7$-$24$ | FC($8$)  & conv$7$-$6$ | conv$7$-$12$ | P | FC(43) \\
                         & CSTN-1 & $5.01\%$ & {[} conv$7$-$6$ | conv$7$-$24$ | FC($8$) {]}$\times 1$  & conv$7$-$6$ | conv$7$-$12$ | P | FC(43)\\
                         & ICSTN-4 & $3.18\%$ & {[} conv$7$-$6$ | conv$7$-$24$ | FC($8$) {]}$\times 4$  & conv$7$-$6$ | conv$7$-$12$ | P | FC(43)\\
                         & CSTN-4 & $3.15\%$ & {[} conv$7$-$6$ | conv$7$-$24$ | FC($8$) {]}$\times 4$  & conv$7$-$6$ | conv$7$-$12$ | P | FC(43)\\ 
                         & \textbf{DeSTNet-4}  & $\mathbf{1.99\%}$ & $\mathcal{F}$\{{[} conv$7$-$6$ | conv$7$-$24$ | FC($8$) {]}$\times 4\}$  & conv$7$-$6$ | conv$7$-$12$ | P | FC(43)\\ \hline \hline
\multirow{6}{*}{\rotatebox[origin=c]{90}{\textbf{MNIST}}}   & CNN & $6.60\%$ & \multicolumn{2}{c}{conv$3$-$3$ | conv$3$-$6$ | P | conv$3$-$9$ | conv$3$-$12$ | FC($48$) | FC($10$)}\\ \cline{2-5} 
                         & STN & $4.94\%$ &conv$7$-$4$ | conv$7$-$8$ | P | FC($48$) | FC($8$)  & conv$9$-$3$ | FC(10)\\
                         & CSTN-1 & $3.69\%$ & {[} conv$7$-$4$ | conv$7$-$8$ | P | FC($48$) | FC($8$) {]}$\times 1$  & conv$9$-$3$ | FC(10)\\
                         & ICSTN-4 & $1.23\%$ & {[} conv$7$-$4$ | conv$7$-$8$ | P | FC($48$) | FC($8$) {]}$\times 4$  & conv$9$-$3$ | FC(10)\\
                         & CSTN-4 & $1.04\%$ & {[} conv$7$-$4$ | conv$7$-$8$ | P | FC($48$) | FC($8$) {]}$\times 4$  & conv$9$-$3$ | FC(10)\\ 
                         & \textbf{DeSTNet-4} & $\mathbf{0.71\%}$ & $\mathcal{F}$\{{[} conv$7$-$4$ | conv$7$-$8$ | P | FC($48$) | FC($8$) {]}$\times 4\}$  & conv$9$-$3$ | FC(10)
\end{tabular}
\end{scriptsize}
\caption{Test classification errors of the compared models on GTSRB and MNIST datasets.}
\label{tbl:ts_mnist_pertrubed}
\end{table}

We compare the performance of DeSTNet to the most related methods,
STN~\cite{jaderberg2015spatial}, a single CSTN (CSTN-1)~\cite{lin2017inverse},
and stack of four CSTNs (CSTN-4)~\cite{lin2017inverse}. For completeness, we report classification
results of a CNN with roughly the same number of learnable parameters and the recurrent version of CSTN (i.e., ICSTN)~\cite{lin2017inverse}. To isolate the contribution of the alignment module, we adopt the same CNN classifier for all. By examining Table~\ref{tbl:ts_mnist_pertrubed}\footnote{\label{symbols_explanation} conv$\text{D}_1$-$\text{D}_2$: convolution layer with
$\text{D}_1\times \text{D}_1$ receptive field and $\text{D}_2$ channels, P: max-pooling layer,
FC: fully connected layer, $\mathcal{F}$: fusion operation used in DeSTNet for combining the
parameters' updates, $\tilde{\mathcal{F}}$: standard fusion operation~\cite{huang2017densely}.} 
we observe that alignment improves classification performance, irrespective of the specific alignment module,
supporting the need for removing perspective transformations with which a standard CNN classifier would not
be able to cope.\footnote{Convolution and max-pooling help with small transformations, but are not enough
to cope with full perspective warpings.} Importantly, CSTN-1 achieves lower classification error as compared
to the STN ($5.01\%$ vs $6.49\%$), thus supporting our architectural choice of building DeSTNet using $\bs{p}$-STNs.
Moreover, using a cascade of four CSTNs further improves results. Finally, the DeSTNet-4 outperforms CSTN-4 
with an error of $1.99\%$ down from $3.15\%$ which amounts to a relative improvement of $~37\%$.
{
\begin{table}[t]
\centering
\begin{scriptsize}
\begin{tabular}{c|c|ccc|c|c}
\multirow{3}{*}{}        & \multirow{3}{*}{\textbf{Model}} & \multicolumn{3}{c|}{\textbf{Test error}} & \multicolumn{2}{c}{\textbf{Architecture}}                        \\ \cline{3-7} 
                         &                        & \multicolumn{3}{c|}{\textbf{Perturbation} $\mathbf{\sigma}$}                            & \multirow{2}{*}{\textbf{Alignment}} & \multirow{2}{*}{\textbf{Classifier}} \\ \cline{3-5}
                         &                        & $\mathbf{10\%}$ & $\mathbf{20\%}$ & $\mathbf{30\%}$             &                            &                             \\ \hline
\multirow{3}{*}{\rotatebox[origin=c]{90}{{\scriptsize \textbf{GTSRB}}}} & CSTN-4                 & $6.86\%$ & $8.92\%$  & $13.72\%$ & {[} conv$7$-$6$ | conv$7$-$24$ | FC($8$) {]}$\times 4$  & FC($43$)\\
                         & \textbf{DeSTNet-4} \textbf{($\mathbf{\tilde{\mathcal{F}}}$)} & $3.60\%$  & $4.65\%$  & $5.25\%$ & $\tilde{\mathcal{F}}$\{{[} conv$7$-$6$ | conv$7$-$24$ | FC($8$) {]}$\times 4\}$ & FC($43$)\\
                         & \textbf{DeSTNet-4} & $\mathbf{3.04\%}$  & $\mathbf{3.80\%}$  & $\mathbf{3.85\%}$ & $\mathcal{F}$\{{[} conv$7$-$6$ | conv$7$-$24$ | FC($8$) {]}$\times 4\}$ & FC($43$)      \\
\hline \hline
\multirow{3}{*}{\rotatebox[origin=c]{90}{{\scriptsize \textbf{MNIST}}}}   & C-STN-4 & $1.50\%$ & $2.39\%$  & $3.40\%$ & {[} conv$7$-$4$ | conv$7$-$8$ | P | FC($48$) | FC($8$) {]}$\times 4$  & FC($10$)\\ 
                         & \textbf{DeSTNet-4} \textbf{($\mathbf{\tilde{\mathcal{F}}}$)} & $0.86\%$  & $0.89\%$  & $1.09\%$ & $\tilde{\mathcal{F}}$\{{[} conv$7$-$4$ | conv$7$-$8$ | P | FC($48$) | FC($8$) {]}$\times 4\}$ & FC($10$)\\
                         & \textbf{DeSTNet-4} & $\mathbf{0.66\%}$  & $\mathbf{0.72\%}$  & $\mathbf{0.74\%}$ & $\mathcal{F}$\{{[} conv$7$-$4$ | conv$7$-$8$ | P | FC($48$) | FC($8$) {]}$\times 4\}$ & FC($10$)                   
\end{tabular}
\end{scriptsize}
\caption{Test classification errors of the compared models by using a single fully connected layer as classifier under three perturbation levels on GTSRB and MNIST datasets.}
\label{tbl:tr_mnist_pertrubed_fcl}
\end{table}
}
\begin{figure}[!b]
\centering
\setlength{\tabcolsep}{2pt}
\begin{tabular}{cccc}
{\small Initial}&\includegraphics[align=c, width=0.28\textwidth]{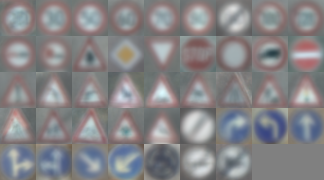} & \includegraphics[align=c, width=0.28\textwidth]{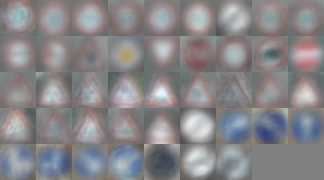}  & 
\includegraphics[align=c, width=0.28\textwidth]{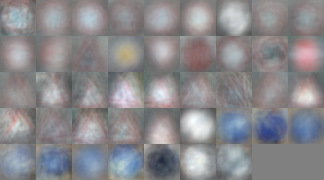} \vspace{0.05cm} \\ 
{\small CSTN-4}& \includegraphics[align=c, width=0.28\textwidth]{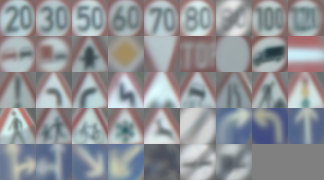} & 
\includegraphics[align=c, width=0.28\textwidth]{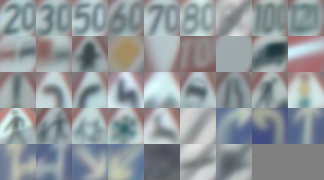}  & 
\includegraphics[align=c, width=0.28\textwidth]{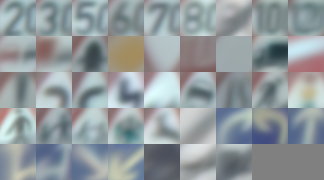} \vspace{0.05cm} \\ 
{\small DeSTNet-4}&\includegraphics[align=c, width=0.28\textwidth]{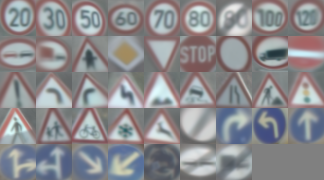} & \includegraphics[align=c, width=0.28\textwidth]{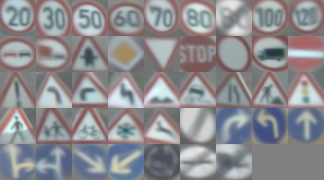}  & 
\includegraphics[align=c, width=0.28\textwidth]{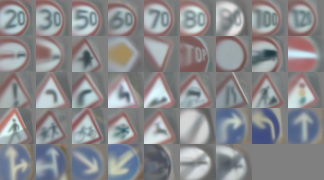} \vspace{0.05cm}\\
& {\small (a) $\sigma=10\%$} & {\small (b) $\sigma=20\%$} & {\small (c) $\sigma=30\%$}
\end{tabular}
\caption{Qualitative comparison of CSTN-4 and DeSTNet-4 methods on GTSRB dataset. Averages of the test traffic signs under different perturbation levels.}
\label{fig:trafic_sign_means}
\end{figure}

\begin{figure}[!t]
\centering
\begin{tabular}{cc}
\includegraphics[width=0.32\textwidth]{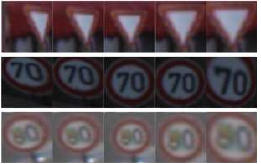} & 
\includegraphics[width=0.32\textwidth]{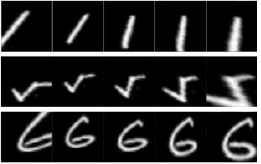} \\
(a) GTSRB & (b) MNIST
\end{tabular}
\caption{Sample alignment results produced by the DeSTNet-4 model on three examples (rows) from the GTSRB (a) and the MNIST (b) datasets. Column 1: input image; columns 2-5: results obtained by applying the intermediate perspective transformations predicted at levels 1-4, respectively.}
\label{fig:aligned_examples}
\end{figure}

It is worth noting, \textit{(i)} the perturbations in this experiment are relatively small
(${\sigma = 10\%}$) and \textit{(ii)} the CNN network followed by a fully connected layer as classifier
does not fully off-load the alignment task to the alignment network. This is due to the translation
invariance and robustness to small transformations brought about by the convolutions and pooling layers.
Therefore, to further investigate the alignment quality of the state-of-the-art CSTN and DeSTNet,
we use a single fully connected layer as a classification network and report performance under three
perturbation levels $\sigma = \{10\%, 20\%, 30\%\}$ corresponding to a minimum of $3.6$ and a maximum
of $10.8$ pixels. Results in Table~\ref{tbl:tr_mnist_pertrubed_fcl}\footnotemark[\getrefnumber{symbols_explanation}]
show that, (\textit{i}) DeSTNet yields an alignment quality that significantly simplifies the classification task compared to CSTN
(i.e., up to $9.87\%$ better classification performance for DeSTNet); (\textit{ii}) DeSTNet exhibits
robustness against stronger perturbation levels, with performance degrading by only $0.81\%$ from 
$10\%$ to $30\%$ perturbation, while CSTN performance degrades by $6.86\%$ in the same range; and 
(\textit{iii}) the proposed \textit{expansion-contraction} fusion block $\mathcal{F}$ leads to better performance w.r.t. 
the standard bottleneck layer $\tilde{\mathcal{F}}$ proposed in~\cite{huang2017densely}. Qualitative experimental results for CSTN and DeSTNet under different perturbation levels are reported in Fig.~\ref{fig:trafic_sign_means}. More specifically, the averages of the $43$ traffic signs before and after convergence for CSTN-4 and DeSTNet-4 are shown. We observe that the average images produced by DeSTNet-4 are much sharper and have more details (even for the $30\%$ perturbation level,
Fig.~\ref{fig:trafic_sign_means}(c)) than the averages produced by CSTN-4, this is indicative of the better alignment performance for the proposed model. Fig.~\ref{fig:aligned_examples}(a) illustrates aligned
examples generated by DeSTNet-4. 
\begin{figure}[!b]
\centering
\setlength{\tabcolsep}{1.5pt}
\begin{tabular}{cccc}
{\small Initial}   & \includegraphics[align=c, width=0.28\textwidth]{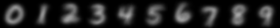}    &  \includegraphics[align=c, width=0.28\textwidth]{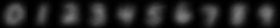}   &   \includegraphics[align=c, width=0.28\textwidth]{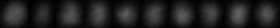}  \\[-2.7pt]

{\small CSTN-4}    & \includegraphics[align=c, width=0.28\textwidth]{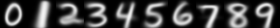}    &  \includegraphics[align=c, width=0.28\textwidth]{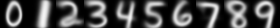}   &   \includegraphics[align=c, width=0.28\textwidth]{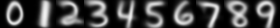}  \\[-2.7pt]

{\small DeSTNet-4} & \includegraphics[align=c, width=0.28\textwidth]{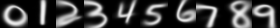}    &  \includegraphics[align=c, width=0.28\textwidth]{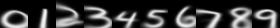}   &   \includegraphics[align=c, width=0.28\textwidth]{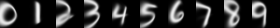}  \\
                          
{\small Initial}   & \includegraphics[align=c, width=0.28\textwidth]{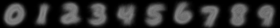}    &  \includegraphics[align=c, width=0.28\textwidth]{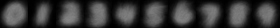}   &   \includegraphics[align=c, width=0.28\textwidth]{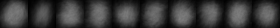}  \\[-2.7pt]
{\small CSTN-4}    & \includegraphics[align=c, width=0.28\textwidth]{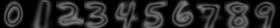}    &  \includegraphics[align=c, width=0.28\textwidth]{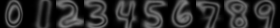}   &   \includegraphics[align=c, width=0.28\textwidth]{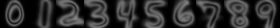}  \\[-2.7pt]

{\small DeSTNet-4} & \includegraphics[align=c, width=0.28\textwidth]{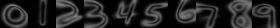}    &  \includegraphics[align=c, width=0.28\textwidth]{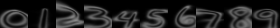}   &   \includegraphics[align=c, width=0.28\textwidth]{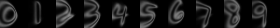}  \\
           & {\small (a) $\sigma=10\%$} & {\small(b) $\sigma=20\%$} & {\small(c) $\sigma=30\%$}
\end{tabular}
\caption{Qualitative comparison of CSTN-4 and DeSTNet-4 on the MNIST dataset. Mean (top rows) and variance (bottom rows) of the $10$ digits under different perturbation levels.}
\label{fig:mnist_res}
\end{figure}
\textbf{Handwritten Digits:} For this experiment, we adopt MNIST dataset~\cite{lecun1998mnist}, consisting of handwritten digits between $0$ and $9$, with a training set of $60,000$ and $10,000$ test grayscale images ($28\times28$ pixels). We adopt the same settings as for the GTSRB experiments by using the image classification error as a \textit{proxy} measure for alignment quality. Training and test sets are distorted using the same perspective warp noise model ($\sigma = 12.5\%$, corresponding to a maximum perturbation of $3.5$ pixels). 

Experimental results are reported in Table~\ref{tbl:ts_mnist_pertrubed}\footnotemark[\getrefnumber{symbols_explanation}]. In line with
the GTSRB experiments, \textit{(i)} pre-alignment considerably improves classification performance,
regardless of the specific alignment module used; \textit{(ii)} lower classification error is achieved when
using CSTN-1 as compared to STN, again supporting our choice of using $\bs{p}$-STNs as base STNs in DeSTNet;
\textit{(iii)} although performance almost saturates with four CSTNs, DeSTNet is still able to squeeze extra
performance, outperforming CSTN-4 with an error of $0.71\%$ down from $1.04\%$ which is a relative
improvement of $32\%$.

We further investigate the alignment quality of the state-of-the-art CSTN and DeSTNet, when a single
fully connected layer is used for classification and report performance under
three perturbation levels corresponding to
a minimum of $2.8$ pixels and a maximum of $8.4$ pixels. By inspecting the results reported in
Table~\ref{tbl:tr_mnist_pertrubed_fcl}\footnotemark[\getrefnumber{symbols_explanation}],
we can see that, \textit{(i)} DeSTNet achieves an alignment quality that significantly simplifies
the classification task compared to CSTN (i.e., up to $2.66\%$ better classification performance for DeSTNet); \textit{(ii)} DeSTNet exhibits robustness against stronger perturbation levels, with the classification performance
degrading by only $0.08\%$ from $10\%$ to $30\%$ perturbation, while CSTN performance degrades by $1.90\%$ in the same range; and \textit{(iii)} the proposed \textit{expansion-contraction} fusion block further helps reducing the classification test error.

Qualitative experimental results are reported in Fig.~\ref{fig:mnist_res}. In particular, the average
and corresponding variance of all test samples grouped by digit are computed and shown for CSTN-4
and DeSTNet-4. Inspecting the images we can see that the mean images generated by DeSTNet-4 are sharper than
those of CSTN-4 while the variance ones are thinner. This suggests that DeSTNet is more accurate and robust to different perturbation levels compared to CSTN. Finally, aligned images generated by the DeSTNet-4 are displayed in Fig.~\ref{fig:aligned_examples}(b). 

\subsection{Document Alignment}
Here, we show how DeSTNet can be successfully utilised for aligning planar images. To this end, we make use of our non-public official identity documents dataset (IDocDB) consisting of $1,000$ training and $500$ testing colour
images collected under in-the-wild conditions. Specifically, each image contains a single identity document
(UK Driving Licence V2015) and their size ranges from $422 \times 215$ to $6,016 \times 3,910$ pixels.
In addition to typical challenges such as non-uniform illumination, shadows, and compression noise, several
other aspects make this dataset challenging, including: the considerable variations in resolution; highly
variable background which may include clutter and non-target objects; occlusion, e.g. the presence of fingers
covering part of the document when held for capture. The ground truth consists of the location of the
four corners of each document. From these points, we can compute a homography matrix that maps each document to a reference frame. The alignment task can be solved by predicting the location of the
corner points on each input image. We train the networks using the smooth $\ell_1$ loss~\cite{ren2015faster}
between the ground truth and the predicted corner coordinates. 
\begin{figure}[t]
\centering
\begin{tabular}{cc}
\includegraphics[width=0.46\textwidth, height=.35\textwidth]{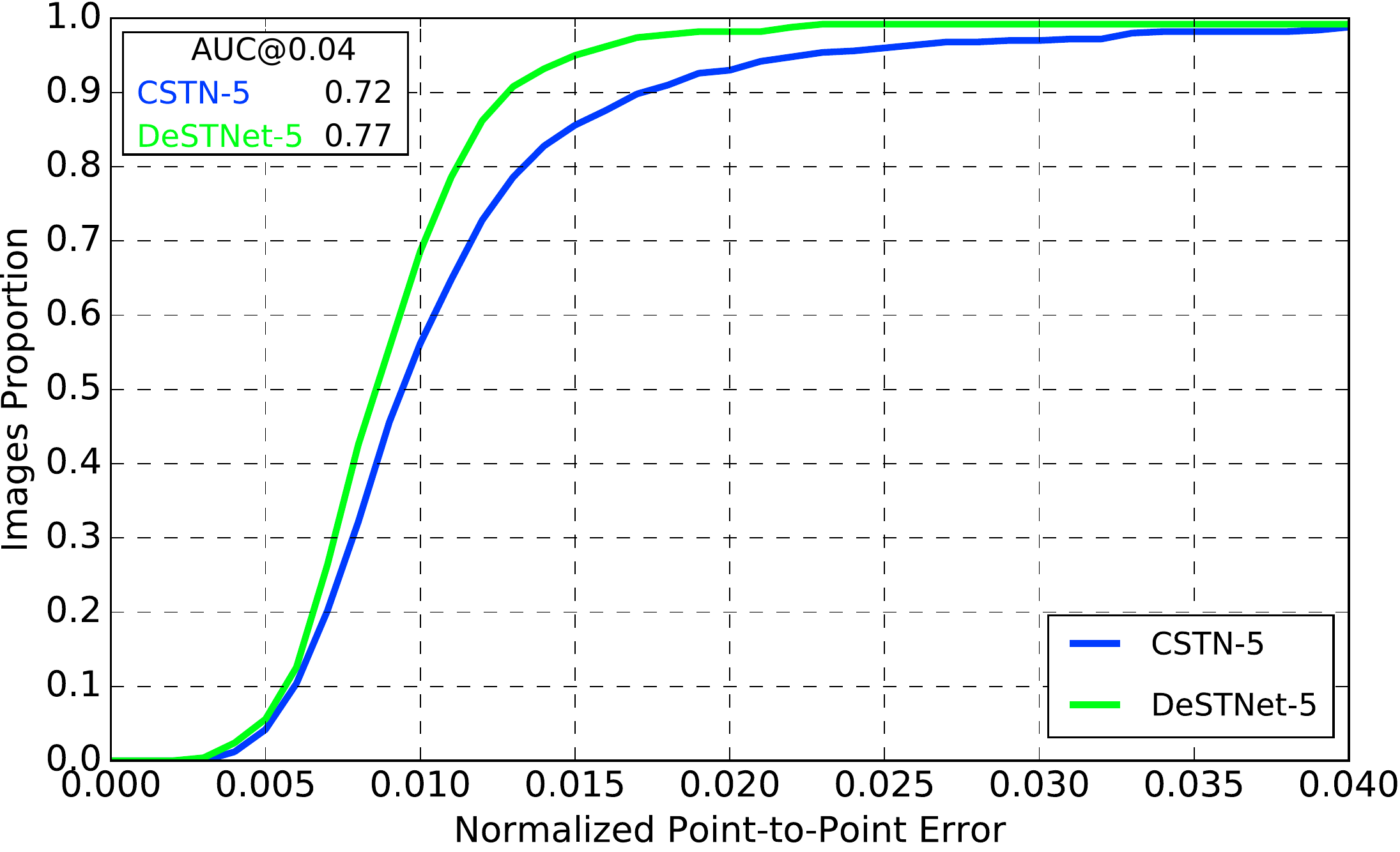} & \includegraphics[width=0.49\textwidth]{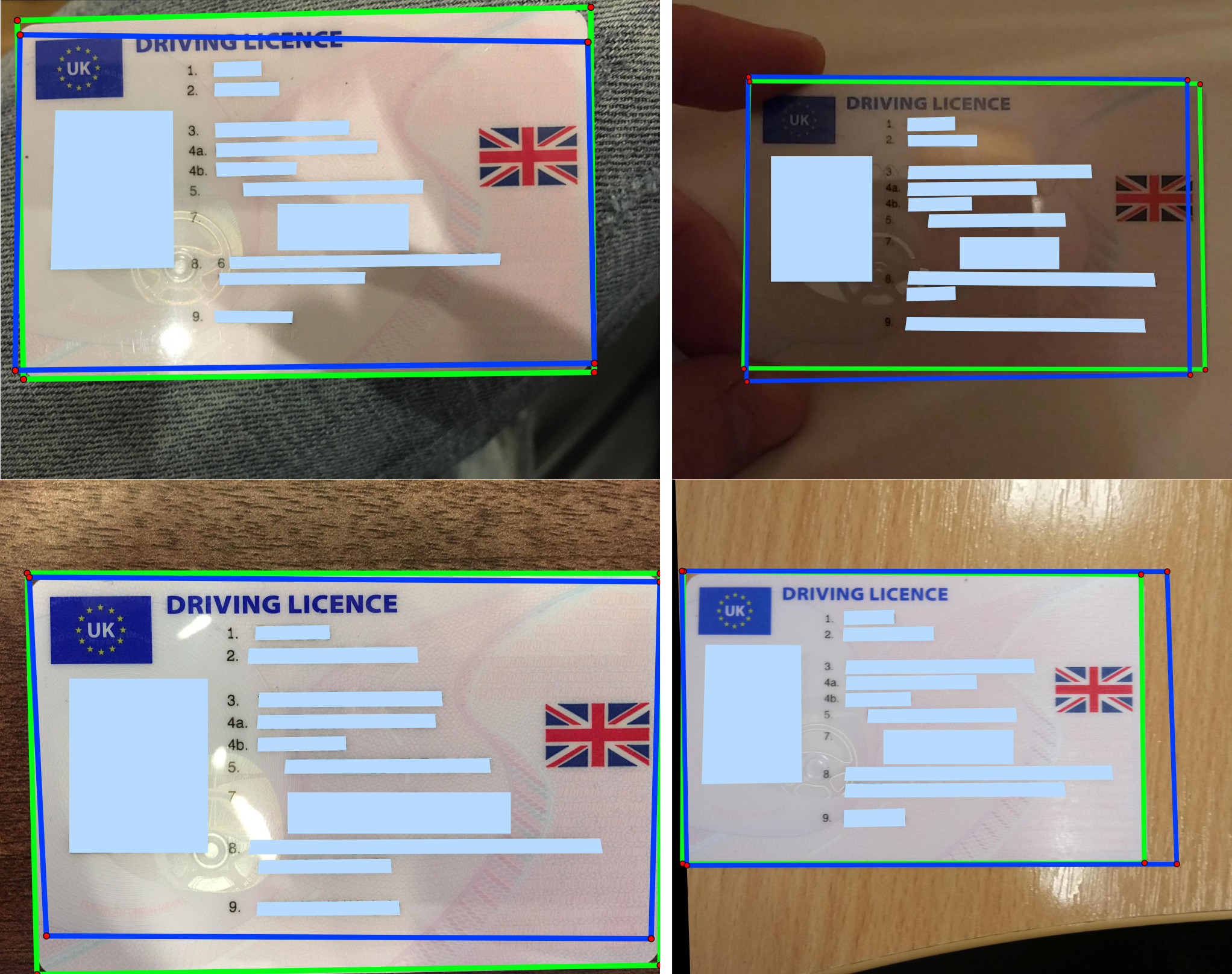} \\
(a) & (b)
\end{tabular}
\caption{(a) Cumulative Error Distribution curves and (b) qualitative results obtained by the {\color{blue} \textbf{CSTN-5}} and {\color{green} \textbf{DeSTNet-5}} on IDocDB.}
\label{fig:docs_roc}
\end{figure}

Adopting the following experimental setting: we resize each image to $256 \times 256$ pixels
for computational efficiency, as done for instance in~\cite{isola2017image,simonyan2014very}. We set the learning rate
for the localisation network to $\alpha_{\text{aln}}=10^{-4}$, which we reduce by $10$ after $20,000$
iterations. We use batches with 8 images each for all the models. For the fusion blocks of DeSTNet,
we set $k^F=256$ and use $S=0.9$. We assess the performance of DeSTNet and compare it with the state-of-the-art CSTNs (strongest baseline based on the presented experiments). Given the increased complexity of the
task compared to MNIST and GTSRB, we built networks with five STNs for both CSTN and DeSTNet
(architectures are reported in Table 1 of supplementary material). For comparison, we use the average point-to-point Euclidean distance, normalised by each
document's diagonal, between the ground truth and predicted location of the four corners. In addition, the Cumulative Error Distribution (CED) curve for each method is computed using the fraction of test images for which the average error is smaller than a threshold. The CED curves in Fig.~\ref{fig:docs_roc}(a) show that DeSTNet-5 outperforms CSTN-5 both in terms of accuracy and robustness. In fact, DeSTNet achieves a higher AUC$@0.04$ ($0.77$ vs $0.72$). Qualitative results for CSTN and DeSTNet are displayed in Fig.~\ref{fig:docs_roc}(b). 
\section{Conclusions}
It is well-known that image recognition is adversely affected by spatial transformations. Increasing geometric invariance helps to improve performance. Although CNNs achieve some level of translation equivariance, they are still susceptible to large spatial transformations. In this paper, we address this problem by introducing DeSTNet, a stack of densely fused STNs that improve information flow in terms of warp parameters' updates. Furthermore, we provide a novel fusion technique demonstrating its improved performance in our problem setting. We show the superiority of DeSTNet over the current state-of-the-art STN and its variant CSTN, by conducting extensive experiments on two widely-used benchmarks (MNIST, GTSRB) and a new non-public real-world dataset of official identity documents.

\textbf{Acknowledgements.} We would like to thank all the members of the Onfido research team for their support and candid discussions.

\bibliography{bmvc_review}

\clearpage

\section{Supplementary Material}
\subsection{Additional Results for Section 4.1}

\begin{figure}[h]
\centering
\includegraphics[width=0.38\textwidth]{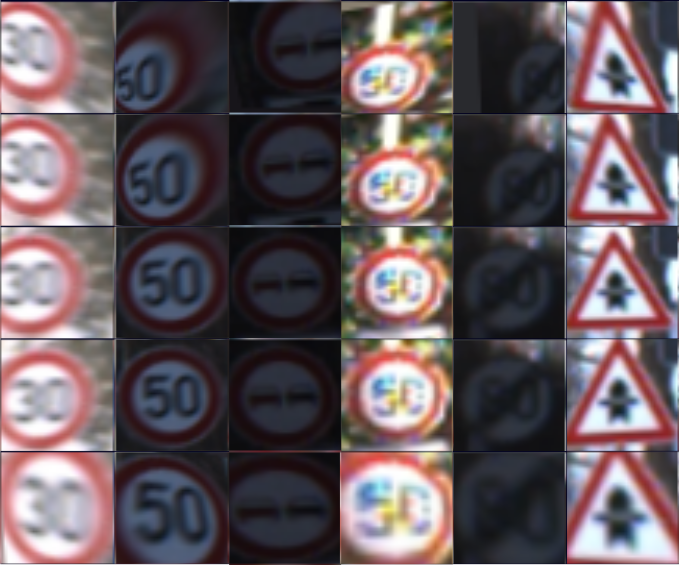} 
\caption{Sample alignment results produced by the DeSTNet-4 model on the GTSRB dataset. Row 1: input image. Rows 2-4: results produced after each one of the four levels.}
\label{fig:tf_samples}
\end{figure}

\begin{figure}[h]
\centering
\includegraphics[width=0.38\textwidth]{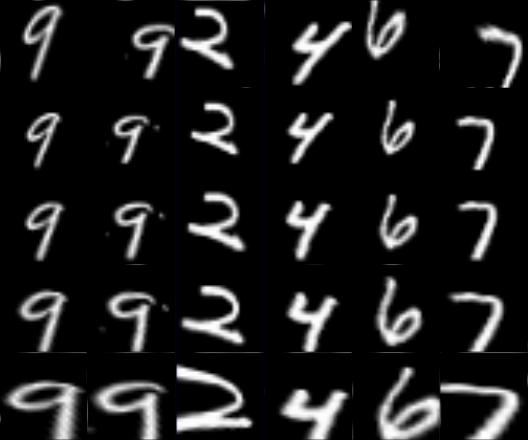} 
\caption{Sample alignment results produced by the DeSTNet-4 model on the MNIST dataset. Row 1: input image. Rows 2-4: results produced after each one of the four levels.}
\label{fig:mnist_samples}
\end{figure}

Figures~\ref{fig:tf_samples} and~\ref{fig:mnist_samples} show additional alignment results obtained by the proposed DeSTNet model on GTSRB~\cite{stallkamp2011german} and MNIST~\cite{lecun1998mnist} datasets, respectively.

~

\subsection{Architectures and Additional Results for Section 4.2}
Table~\ref{tbl:archs} reports the architectures of the compared CSTN-5~\cite{lin2017inverse} and DeSTNet-5 models for the task of planar image alignment.

\begin{table}
\centering
\begin{tabular}{c|c}
\textbf{Model} & \textbf{Architecture} \\\hline
CSTN-5 & {[} conv$3$-$64(2)$ | conv$3$-$128(2)$ | conv$3$-$256(2)$ | FC$8$ {]}$\times 5$\\
\textbf{DeSTNet-5} & $\mathcal{F}$\{{[} conv$3$-$64(2)$ | conv$3$-$128(2)$ | conv$3$-$256(2)$ | FC$8$ {]}$\times 5\}$
\end{tabular}
\caption{Architectures utilized by CSTN-5 and DeSTNet-5. conv$\text{D}_1$-$\text{D}_2(\text{D}_3$): convolution layer with $\text{D}_1\times \text{D}_1$ receptive field, $\text{D}_2$ channels and $\text{D}_3$ stride, FC: fully connected layer, $\mathcal{F}$: fusion operation used in DeSTNet for fusing the parameters updates.}
\label{tbl:archs}
\end{table}

\begin{figure}[!t]
\centering
\begin{tabular}{c}
\includegraphics[width=1.0\textwidth]{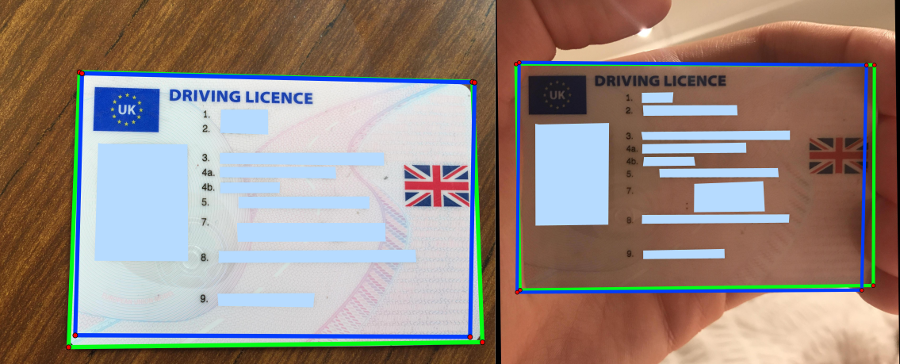} \\
\includegraphics[width=1.0\textwidth]{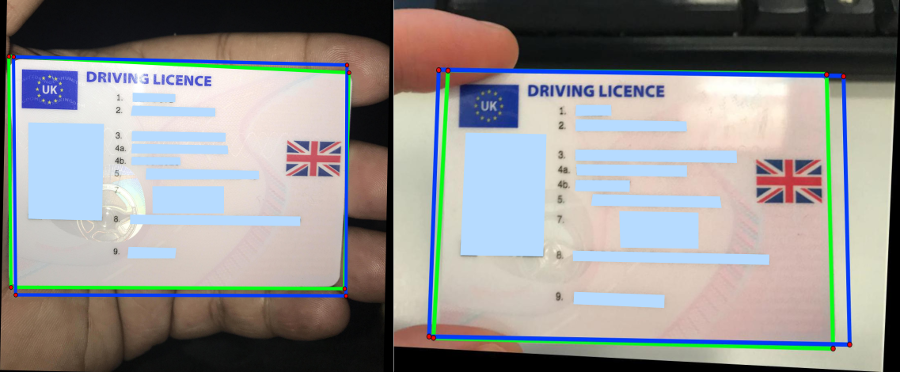} \\
\includegraphics[width=1.0\textwidth]{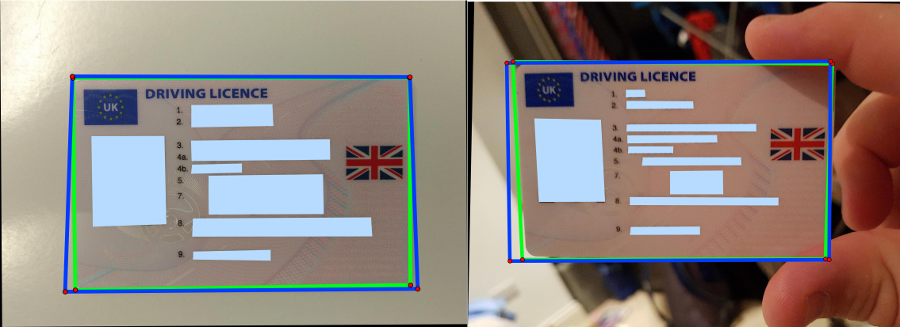} 
\end{tabular}
\caption{Qualitative results obtained with {\color{blue} \textbf{CSTN-5}} and {\color{green}
\textbf{DeSTNet-5}} on IDocDB. (Results are best viewed on a digital screen)}
\label{fig:doc_samples_1}
\end{figure}

\begin{figure}[!t]
\centering
\begin{tabular}{c}
\includegraphics[width=1.0\textwidth]{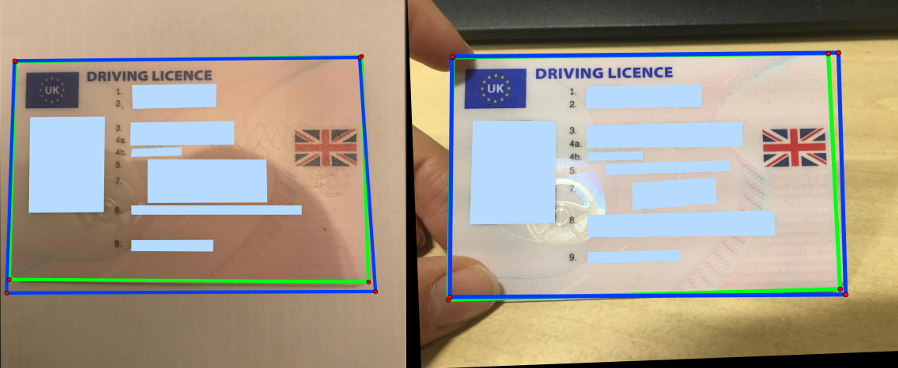} \\
\includegraphics[width=1.0\textwidth]{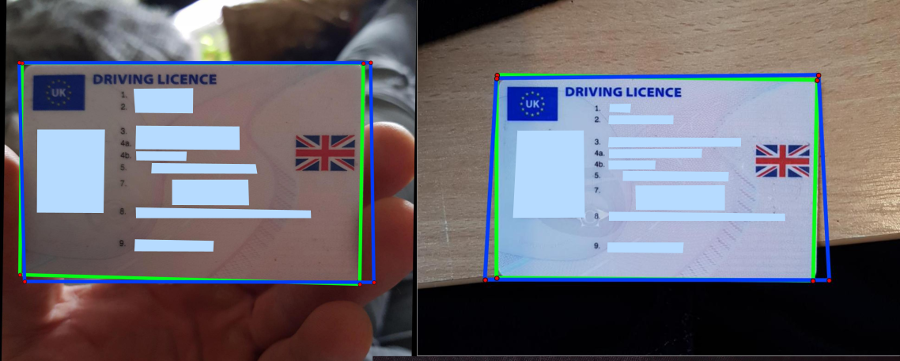} \\
\includegraphics[width=1.0\textwidth]{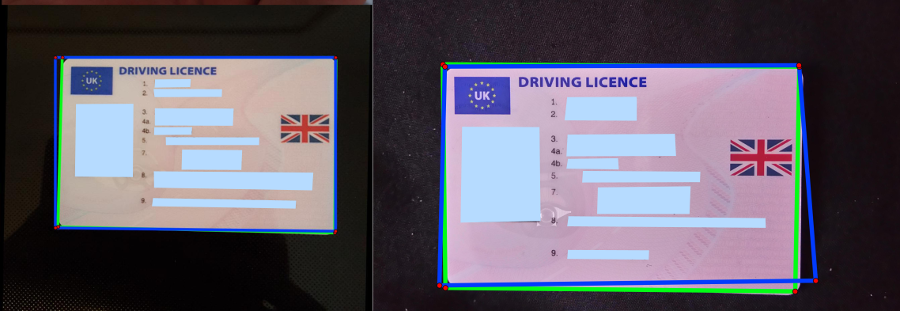} 
\end{tabular}
\caption{Qualitative results obtained with {\color{blue} \textbf{CSTN-5}} and {\color{green}
\textbf{DeSTNet-5}} on IDocDB. (Results are best viewed on a digital screen)}
\label{fig:doc_samples_2}
\end{figure}

%
Additional qualitative results obtained by the CSTN-5 and DeSTNet-5 on the IDocDB database are provided in Figs.~\ref{fig:doc_samples_1},~\ref{fig:doc_samples_2}. These results confirm that the proposed DeSTNet is more accurate than the CSTN and show better robustness against partial-occlusions, clutter and low-light conditions.

\end{document}